\documentclass[10pt,twocolumn,letterpaper]{article}

\usepackage[pagenumbers]{cvpr} 

\usepackage{graphicx}
\usepackage{amsmath}
\usepackage{amssymb}
\usepackage{booktabs}
\usepackage{verbatim}
\usepackage{lib}
\usepackage{cuted}
\usepackage{capt-of}
\usepackage{booktabs}
\usepackage{caption}
\usepackage{stfloats}
\usepackage{graphicx}
\usepackage{placeins}
\usepackage{axessibility}

\usepackage[pagebackref,breaklinks,colorlinks]{hyperref}

\usepackage[capitalize]{cleveref}
\crefname{section}{Sec.}{Secs.}
\Crefname{section}{Section}{Sections}
\Crefname{table}{Table}{Tables}
\crefname{table}{Tab.}{Tabs.}

\def\cvprPaperID{4391} 
\def\confName{CVPR}
\def\confYear{2022}

\newcommand\shelfbl{SSMP~\cite{ye2021shelf}\xspace}
\newcommand\syntheticbl{Synth.\xspace} 
\newcommand\singlebl{Cat-spec.\xspace} 
\newcommand\distillbl{Ours\xspace} 

\begin{document}

\title{Pre-train, Self-train, Distill: A simple recipe for Supersizing 3D Reconstruction}

\author{Kalyan Vasudev Alwala\textsuperscript{1}  \qquad Abhinav Gupta\textsuperscript{12} \qquad Shubham Tulsiani\textsuperscript{2}   \\
\textsuperscript{1}Meta AI Research \qquad \textsuperscript{2}Carnegie Mellon University \\

{\tt \small \href{https://shubhtuls.github.io/ss3d/}{https://shubhtuls.github.io/ss3d/}}
}

\maketitle
 \begin{strip}\centering
\vspace{-1.5cm}
\includegraphics[width=\textwidth]{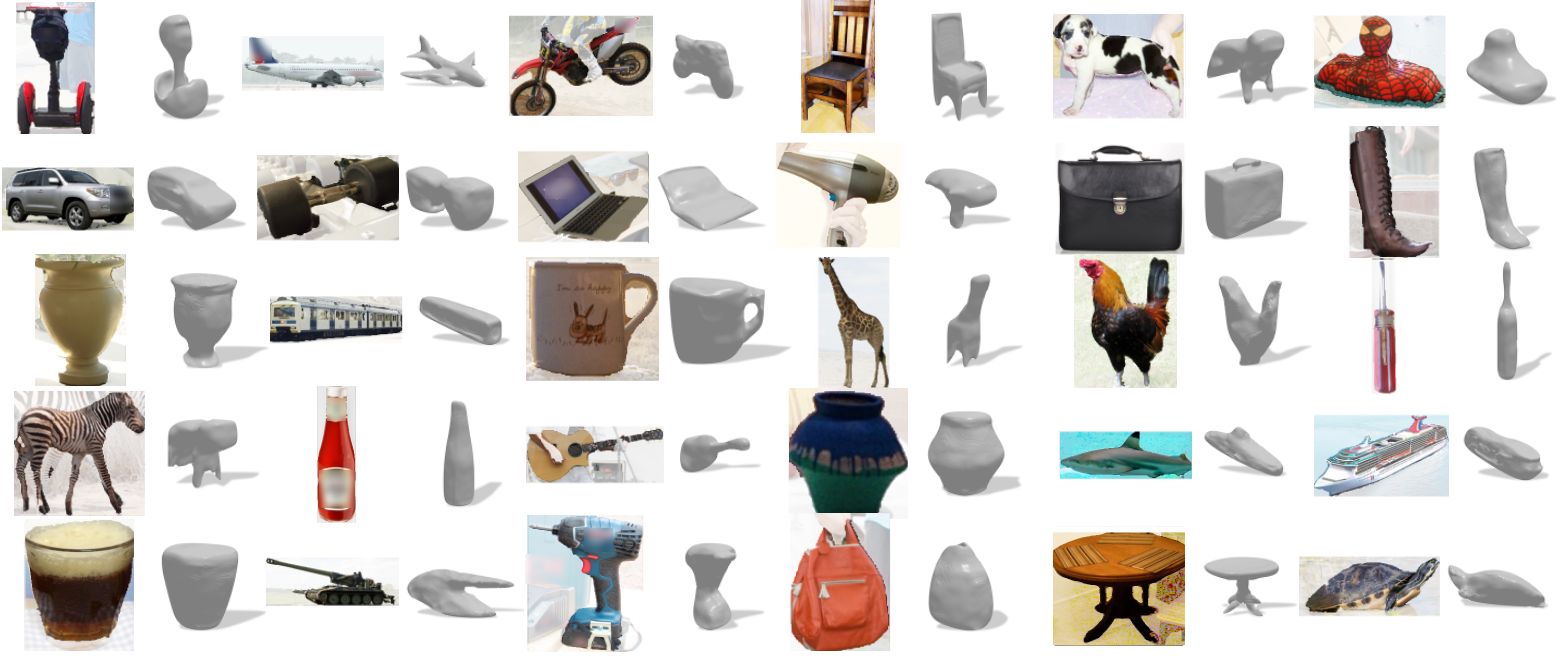}
\captionof{figure}{
We present an approach to learn a \emph{single} self-supervised reconstruction model across diverse object categories. Given an input image depicting any (segmented) object spanning over 150 categories, this unified reconstruction model can infer its 3D shape.}
\label{fig:teaser}
\vspace{-.2cm}
\end{strip}

\begin{abstract}
\vspace{-4mm}
    Our work learns a unified model for single-view 3D reconstruction of objects from hundreds of semantic categories. As a scalable alternative to direct 3D supervision, our work relies on segmented image collections for learning 3D of generic categories. Unlike prior works that use similar supervision but learn independent category-specific models from scratch, our approach of learning a unified model simplifies the training process while also allowing the model to benefit from the common structure across categories. Using image collections from standard recognition datasets, we show that our approach allows learning 3D inference for over 150 object categories. We evaluate using two datasets and qualitatively and quantitatively show that our unified reconstruction approach improves over prior category-specific reconstruction baselines. Our final 3D reconstruction model is also capable of zero-shot inference on images from unseen object categories and we empirically show that increasing the number of training categories improves the reconstruction quality.
\end{abstract}

\section{Introduction}
We live in a rich, diverse world comprising of a variety of objects, from naturally occurring birds and bears, to ingeniously engineered cars and airplanes, or the pragmatically simple cups and vases. While we have made significant advances in semantic understanding of the
visual world, developing approaches which can recognize hundreds and thousands of these objects, the state-of-the-art single-view reconstruction systems are only capable of inferring 3D for a handful among them. Why is that? 

\begin{figure*}[t!]
    \centering
    \includegraphics[width=0.9\textwidth]{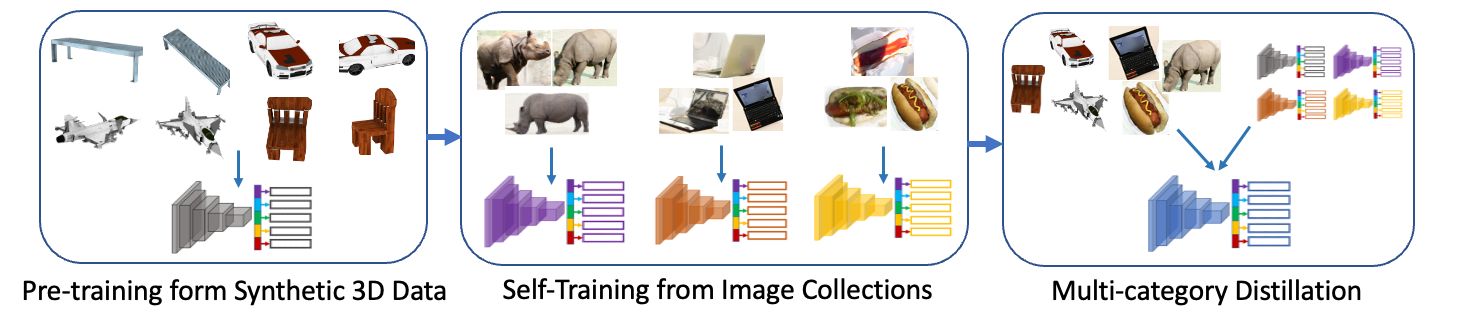}
    \vspace{-0.1cm}
    \caption{\textbf{Approach Overview.} We first pre-train a reconstruction model using multi-view renderings of synthetic data. We then self-train category-specific models on diverse image collections in-the-wild with only foreground mask annotations. We finally distill the learned models from prior training stages into a unified reconstruction model.}
    \label{fig:overview}
\end{figure*}

The primary reason for success in case of semantic recognition is the use of supervision. Availability of large amounts of training data for thousands of categories for classification~\cite{deng2009imagenet}, detection~\cite{kuznetsova2020open} and even segmentation~\cite{gupta2019lvis} has been the key to progress. On the other hand, obtaining 3D supervision  for images of generic objects is extremely hard. Even the biggest 3D datasets contain only tens of categories and even in that case, the images do not reflect the complexity of the real-world. 

To tackle this problem of supervision, a number of recent works have instead investigated learning 3D inference using only image collections and 2D annotations (such as foreground masks) as supervisory signal. However, as these approaches learn category-level models from scratch, they have no 3D priors and need to leverage additional constraints or regularization to make learning tractable. Unfortunately, this either restricts their applicability to generic classes \eg commonly used morphable models cannot capture categories with large variation, or makes the training process unstable \eg when using adversarial constraints. 

Most approaches in the area have adopted one of the two extremes: supervised or self-supervised and therefore compromised either on scalability or generalization. 
In this paper, instead of adopting two extreme viewpoints -- we adopt the middle ground. More specifically, we adopt three key strategies to achieve both scalability and tractability of learning 3D reconstruction. First, we observe that while semantic recognition approaches train one model to detect hundreds of categories; single-view reconstruction approaches train one model for each category. This bottlenecks the sharing of primitives and features across multiple categories. We argue this sharing is critical for generalization. In this work, we propose to train a \emph{single} model to reconstruct objects across hundreds of diverse object categories. Second, instead of learning this joint model via regularization or extra constraints, we use synthetic 3D training data for \emph{some} categories to help us learn priors. More specifically, we pre-train a base model using the synthetic data in a supervised manner. Third and finally, we observe that a pre-trained model cannot be used to just fine-tune on all the new categories in one shot. This is because mask-supervision is weak supervision at best and joint learning leads to averaging errors. Instead, we propose a self-training label propagation procedure: first fit a category specific model followed by a distillation process to learn a joint model that provides better generalization. 

Our final approach is combination of pre-training and self-training that provides robustness and scalablilty. We first pretrain a 3D reconstruction model using the 3D synthetic data. This pre-training stage helps the model learn right 3D priors to learn from weak mask supervision. One we learn.a pre-trained base model, we then adapt this base model trained on synthetic data to reconstruct novel categories from image collections, and show that this obviates the need for additional regularization such as limited intra-class variation or adversarial priors. In the final step, we then distill the per-category adapted models into a unified high-capacity 3D reconstruction network which is finally able to reconstruct objects from across hundreds of object categories (see Fig.\ref{fig:teaser}). Interestingly, not only can this unified model better reconstruct the categories seen during training, it can even be used for \emph{zero-shot} 3D prediction for objects from \emph{unseen} categories -- something that prior  per-category learning methods are fundamentally incapable of.

\section{Related Work}


\vspace{-2mm}
\paragraph{Learning 3D from Supervision.} 
With the resurgence of neural networks, several works have leveraged deep learning techniques for single-view 3D reconstruction. Using ground-truth 3D as supervision, prior approaches have pursued inference of representations such as voxels~\cite{girdhar2016learning,choy20163d} , meshes~\cite{wang2018pixel2mesh}, Point Clouds ~\cite{fan2017point}, octrees ~\cite{hane2017hierarchical} or implicit functions ~\cite{mescheder2019occupancy, park2019deepsdf}. However, due their reliance on 3D ground truth data, these methods mostly rely on synthetic training data and are limited in their scalability to generic object categories. To overcome this reliance on 3D supervision, subsequent approaches~\cite{liu2019learning,wiles2017silnet,tulsiani2017multi} instead used multi-view image collections to provide supervisory signal. However, as  multi-view data is also difficult to acquire in-the-wild, other methods proposed to learn 3D from single-view images, although relying on additional annotations such as camera poses~\cite{lin2020sdf, henderson2020learning,henderson2020leveraging, kato2019learning} or semantic keypoints~\cite{kong2019deep,kanazawa2018learning}. While these works have all demonstrated impressive results, they critically require some form of annotations (ground truth 3D,  multiple views, 2D keypoints, or camera poses) that are hard to acquire for generic classes. While our approach similarly relies on synthetic 3D data for a base set of classes to bootstrap learning, we show that 3D inference for \emph{novel} categories can be learned without any such annotations, and thus allows our method to scale to hundreds of object classes using in-the-wild image collections.

\begin{table}
\centering

\resizebox{\columnwidth}{!}{
\begin{tabular}{@{}r|rrrrrrrrr@{}  }

  &  SRN-SDF  & CMR & U-CMR & IMR & UMR  & SSMP  & Ours\\
  &   \cite{lin2020sdf} &\cite{kanazawa2018learning} & \cite{ucmrGoel20} & \cite{tulsiani2020implicit} & \cite{li2020self}   & \cite{ye2021shelf}  & \\
  \hline
  camera                      & \checkmark&  \checkmark& & & & & \\
  template                    & &  \checkmark&\checkmark & \checkmark& & & \\
  keypoints                   & &  \checkmark& & & & & \\
  mask                        &\checkmark &  \checkmark&\checkmark &\checkmark & \checkmark & \checkmark& \checkmark\\
  \hline
  Per-Category Model          & & \checkmark &\checkmark &\checkmark &\checkmark &\checkmark & \\
  Number of Categories        & $<$15  &  $<$5  &$<$15 &$<$25 & $<$10 & $<$75 & $>$150\\
  Shape-Regularizers          & & Def & Def & Def& Adv & Adv & \\
  \hline  
\end{tabular}}
\vspace{-2mm}
\caption{\label{tab:related} A summary of the requirements and limitations of existing single-view 3D reconstruction approaches. Def and Adv denote morphable model and adversarial regularizers respectively.}
\end{table}
\vspace{-5mm}

\paragraph{Learning 3D from Unannotated Image Collections.}
Closer to our goal of scalably learning 3D prediction,  recent works have shown that single-view 3D inference can be learned from widely available category-level image collections, requiring only foreground masks as additional supervision. All these methods, whose training and inference setups we summarize in Table.\ref{tab:related}, learn 3D prediction by enforcing reprojection consistency with the available observations. While this removes the dependency on expensive annotations, these approaches require additional constraints to avoid degenerate solutions. For example, one common solution is to enforce that the variation within a category can be captured by a linear morphable model~\cite{ucmrGoel20,tulsiani2020implicit,li2020self} and optionally use category-level templates \cite{ucmrGoel20, tulsiani2020implicit,kulkarni2020articulation,kulkarni2019canonical} or self-supervised semantics~\cite{li2020self} to resolve pose ambiguities. Unfortunately, this prevents these approaches from reconstructing categories with non-spherical topology \eg handbags or significant variation \eg chairs. An alternative approach is
to allow more expressive shape models and rely on adversarial training to encourage novel-view renderings to be realistic~\cite{nguyen2019hologan,ye2021shelf,henzler2019escaping}. However, these methods can be more difficult to tune and also require a prior over training viewpoints. Our key insight is that these adhoc regularizers are required because all these methods learn per-category models from scratch. In contrast, we propose a simple and scalable approach for learning 3D that leverages synthetic data for pretraining and learns a unified model across object categories, thus allowing us to learn 3D for novel categories without requiring any such explicit regularizers.

\vspace{-2mm}
\paragraph{Implicit Shape Representations.} Approaches tackling unsupervised 3D reconstruction typically infer meshes~\cite{ucmrGoel20,li2020self,tulsiani2020implicit} or discrete volumes~\cite{nguyen2019hologan,ye2021shelf}, but these are limited in their expressiveness or require learned neural renderers to model appearance~\cite{nguyen2019hologan}. Inspired by their suitability for volume rendering and  success in modeling complex scenes~\cite{mildenhall2020nerf}, we instead opt to use neural implicit functions to represent the 3D shape (and appearance). Although the more striking recent applications of these representations have been to model single instances~\cite{sitzmann2020implicit,mildenhall2020nerf,yariv2020multiview}, our approach requires modeling \emph{different} objects via a single network, and we therefore use image-conditioned implicit networks. Prior works have proposed latent variable-based~\cite{mescheder2019occupancy,park2019deepsdf}, hyper-network based~\cite{sitzmann2019scene} or pixel-aligned~\cite{yu2021pixelnerf} mechanisms for such conditioning, but we instead adapt a modulation mechanism commonly used in generative networks~\cite{chan2021pi,karras2019style}. While the choice of our representation is shared with these recent approaches, our work shows that these representations can be learned from unannotated image collections, and that a single conditional implicit network can capture 3D across several categories.


\section{Approach}

\begin{figure}[t!]
    \centering
    \includegraphics[width=0.9\linewidth]{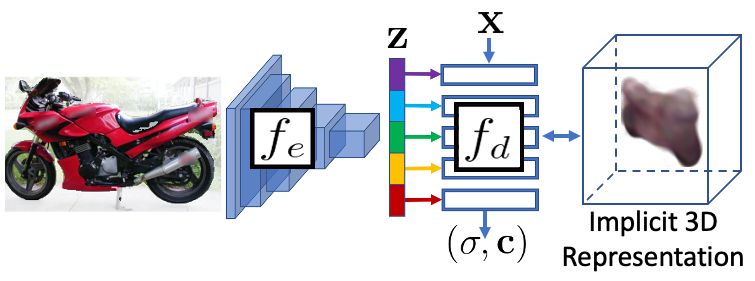}
    \vspace{-0.1cm}
    \caption{\textbf{Network Architecture.} Our image-conditioned implicit reconstruction network uses a ResNet-based encoder. The predicted encoding is used to conditionally modulate the outputs of the intermediate layers of a coordinate-based implicit network.}
    
    \figlabel{architecture}
\end{figure}

\begin{figure*}[t!]
    \centering
    \includegraphics[width=0.9\textwidth]{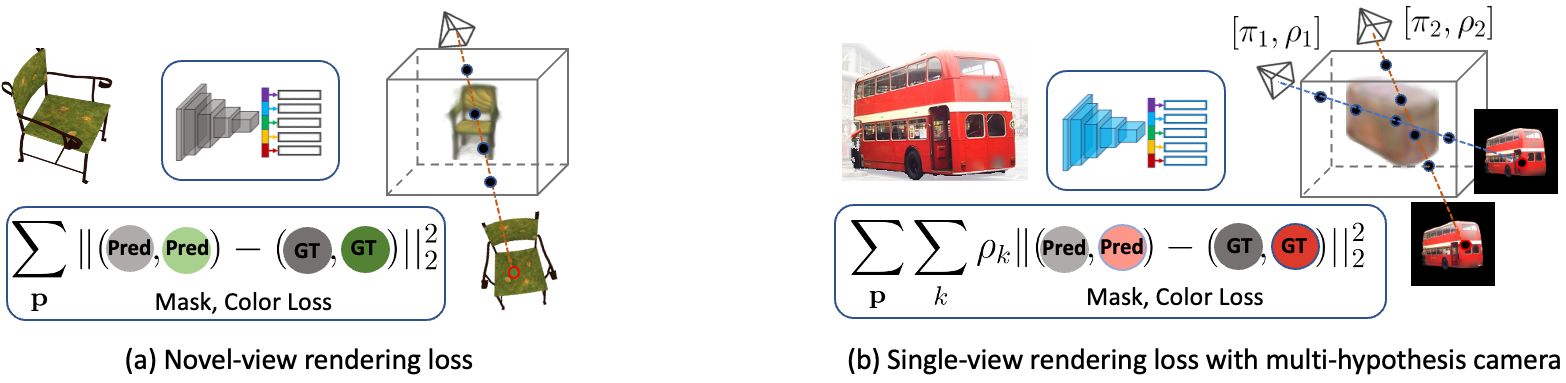}
    \vspace{-0.1cm}
    \caption{\textbf{Training Objectives.}  \textbf{left:} When pre-training with synthetic data, we supervise the reconstruction network via a novel-view rendering loss. For each pixel/ray in a novel view, we volume-render the predicted mask and color using the implicit representation and penalize deviations from the ground-truth. \textbf{right:} To self-train a category-level expert from image collections, we minimize the expected rendering loss under a multi-hypothesis camera parametrization.}
    \figlabel{approach}
    
\end{figure*}

\begin{figure}[t!]
    \centering
    \includegraphics[width=\linewidth]{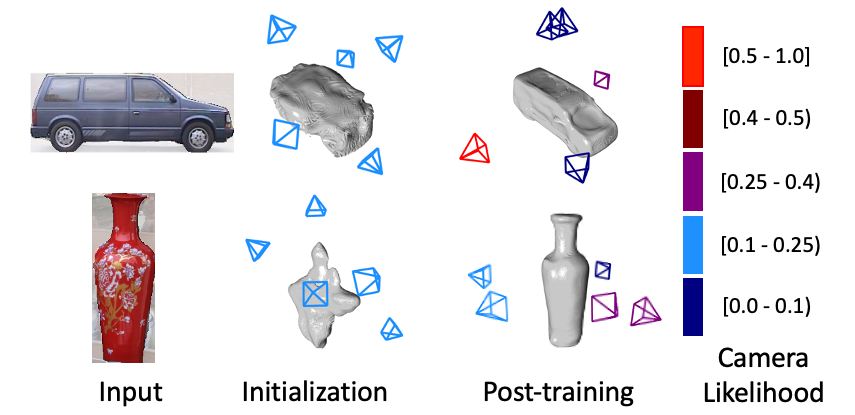}
    \caption{\textbf{Shape and Camera optimization in Self-training.}  We visualize the initial and final configurations for two images. Although the symmetric vase maintains a multi-modal camera distribution, the shape for both instances improves with training.}
    \vspace{-3mm}
    \label{fig:cam_multiplex}
\end{figure}


Our goal in this work is to learn a unified model that can infer 3D from a single image of any object from over hundreds of categories. As 3D supervision is naturally difficult to obtain for such diverse set of categories, our approach relies on single-view image collections with approximate foreground masks as a more scalable alternative. However, instead of solely relying on these image collections, our approach is driven by the insight that (synthetic) 3D data for \emph{some} categories does exist. While image collections are perhaps the only scalable source for learning about categories like balloons, or bananas, or starfishes, we do have 3D data available for other categories like cars and chairs. To incorporate this observation, we propose a multi-stage training approach that leverages both, synthetic shape collections and in-the-wild image collections for learning 3D inference across a broad set of categories.

In the first stage of training (\secref{synthetic_approach}), we pretrain our image-conditioned implicit reconstruction network using synthetic textured 3D models for about 50 object categories. The availability of 3D models allows us to render multiple views of the same instance while also knowing the precise camera poses for each image, thus allowing us to train using a novel-view rendering objective. However, as this initial model is only trained with synthetic renderings on a limited set of categories, it unsurprisingly does not perform well in-the-wild. In the second stage of learning (\secref{singlecat_approach}), we finetune this initial model using in-the-wild category-level image collections. Although neither camera viewpoints nor multiple views are available for learning, we find that the generic 3D priors learned in the first stage help prevent degenerate solutions. This yields multiple category-specific models, and we propose a simple technique to then distill these into a unified reconstruction model which can be used to perform 3D inference across generic classes (\secref{distillation_approach}).

\vspace{-2mm}
\paragraph{Image-conditioned Implicit Reconstruction Network.} Our reconstruction model(s) can be viewed as having an encoder-decoder structure.  Given an input image $I$, the encoder $f_e: I \rightarrow \mathbb{R}^d$ maps it to a latent code $\mathbf{z} = f_e(I)$. Our implicit decoder uses this latent code as a conditioning to predict the density $\sigma \in \mathbb{R}^1$ and color $\mathbf{c} \in \mathbb{R}^3$ for any query point $\mathbf{x}$ \ie $f_d(\mathbf{x}, \mathbf{z}) = (\sigma, \mathbf{c})$. The image-conditioned decoder $f_d(\cdot, \mathbf{z})$ can thus be viewed as a implicitly representing the predicted geometry and appearance for the object depicted in the input image, and can be volume-rendered to generate images from any query viewpoint.

Our network architecture is highlighted in \figref{architecture}. The encoder is based on a ResNet-34~\cite{he2016deep} architecture and outputs a high-dimensional latent code given the input image. Inspired by Chan \etal~\cite{chan2021pi} who used a similar architecture for unconditionally generating radiance fields, we use a SiREN~\cite{sitzmann2020implicit} based decoder network with FiLM conditioning~\cite{perez2018film}, where the  parts of the latent code modulate the output of each decoder layer. Note that we do not  condition the reconstruction on any explicit category labels, and the network is tasked to infer 3D for generic objects given only the (segmented) input image.

\vspace{-2mm}
\paragraph{Volume Rendering.} We denote by $\mathcal{V}(f, \pi, \mathbf{p})$ the process of volume rendering an implicit function $f$ over the ray corresponding to a pixel $\mathbf{p}$ in an image taken from viewpoint $\pi$. Given a query ray, we follow the approach in NeRF~\cite{mildenhall2020nerf} for differentiable volume rendering~\cite{drebin1988volume} with a neural representation by uniformly sampling 3D points along the ray, and aggregating the outputs using the predicted densities (and colors) to render. We use $\mathcal{V}_{m}$ and $\mathcal{V}_{c}$ to denote mask and color rendering respectively, and note the mask rendering process is equivalent to using a constant unit color at each 3D point. We refer the reader to ~\cite{mildenhall2020nerf} for more details.

\subsection{Pre-Training from Synthetic 3D Data}
\seclabel{synthetic_approach}

We pretrain our implicit reconstruction  network using a dataset of synthetic 3D shapes. As we have access to textured 3D meshes, we are able to render multiple views for each object with known camera viewpoints. We then train our reconstruction network by simply enforcing that the single-view (implicit) 3D predictions, when volumetrically rendered from the available viewpoints, match the known color and masks at each pixel.

Given an input image $I$ depicting a \emph{segmented} object, our network predicts an implicit 3D representation $f_I \equiv f_d(\cdot, f_e(I))$. We then use an image $\bar{I}$ of the same instance, captured from a novel camera viewpoint $\bar{\pi}$ to supervised our prediction. Denoting by $\bar{I}_{m}[\mathbf{p}]$ and $\bar{I}_{c}[\mathbf{p}]$ the observed color and foreground mask label for a pixel $\mathbf{p}$, and using $\{m,c\}$ as shorthand for indexing over mask and color, we train our network to minimize the reconstruction error between the predictions and the renderings across the dataset:
\begin{equation*}
\label{eq:synthetic_loss}
 L_{\text{synth}} = \sum_{\mathbf{p}}~ \|\mathcal{V}_{\{m,c\}}(f_I, \bar{\pi}, \mathbf{p}) - \bar{I}_{{\{m,c\}}}[\mathbf{p}] \|^2_2
\end{equation*}

\subsection{Self-Training from Image Collections}
\seclabel{singlecat_approach}
\vspace{-1mm}
Our approach aims to adapt the pretrained implicit reconstruction network to generic novel categories. While the methodology described in \secref{synthetic_approach} allows learning single-view 3D prediction, it crucially relies on available multi-view renderings and associated camera poses. However, when learning 3D reconstruction for generic categories in-the-wild, this form of supervision is not available. Instead, we must rely on single-view image collections with approximate foreground masks for training. Our approach is to derive supervisory signal by enforcing consistency between renderings of the predicted 3D and the input image, but we need to overcome the challenges posed by unknown camera viewpoints in order to do so.

\vspace{-2mm}
\paragraph{Volume Rendering with Multi-hypothesis Cameras.} When jointly learning (unknown) shape and camera viewpoints with differentiable rendering, the optimization is susceptible to local minima. To make the optimization landscape easier, several prior works~\cite{tulsiani2018multi,insafutdinov2018unsupervised,kulkarni2019canonical,ucmrGoel20} have advocated using a \emph{multi-hypothesis} viewpoint parametrization $\Pi \equiv \{(\pi_1, \rho_1),\dots,(\pi_K,\rho_K)\}$, with a probability $\rho_k$ associated with hypothesis $\pi_k$. Under this probabilistic parametrization for camera viewpoint, a rendering loss can be formalized as an \emph{expected} loss -- with the loss under each hypothesis weighted by its likelihood. Given an image $I$, and camera $\Pi$, the volume rendering loss w.r.t. an implicit representation $f$ can be computed as:
\begin{equation*}
L(f, I, \Pi) = \sum_{\mathbf{p}} \sum_{k}  \rho_k \|\mathcal{V}_{\{m,c\}}(f, \pi_k, \mathbf{p}) - I_{{\{m,c\}}}[\mathbf{p}] \|^2_2
\end{equation*}
In practice, as volume rendering is computationally expensive, we approximate this loss by only sampling a small number of pixels/rays for each camera hypothesis.

\vspace{-2mm}
\paragraph{Finetuning 3D Inference in-the-wild.} Given a category-level collection of images with associated foreground masks $\mathcal{D} \equiv \{(I^n, I^n_m)\}$, we adapt the implicit reconstruction network learned in \secref{synthetic_approach} to reconstruct instances in this collection. Inspired by work from Goel \etal~\cite{ucmrGoel20}, we first associate each image $I^n$ with a randomly initialized multi-hypothesis camera $\Pi^n$. We then formulate our learning problem as that of jointly optimizing the 3D reconstruction network and the per-image camera hypotheses $\Pi^n$, and minimize the following objective:
\begin{equation*}
    L_{\text{img}} = \sum_n L(f_d(\cdot, f_e(I^n)), I^n, \Pi^n)
\end{equation*}
Similar to prior approaches using multi-hypothesis cameras~\cite{kulkarni2019canonical,ucmrGoel20}, we follow a two-step optimization approach where only the camera parameters are initially learned while keeping the 3D model fixed. Unlike U-CMR~\cite{ucmrGoel20}, we also optimize the probabilities $\rho^n_k$ instead of treating these as deterministic functions of the rendering loss. A more fundamental difference with the prior approaches~\cite{kulkarni2019canonical,ucmrGoel20} though is that we do not rely on a template 3D shape to make the camera optimization well-behaved. Instead, we find the shapes inferred by our pretrained reconstruction network are sufficient to guide initial camera optimization. After this initial optimization, we then jointly finetune the reconstruction network while continuing to optimize the cameras  $\{\Pi^n\}$ over the image collection.

\begin{figure}[t!]
    \centering
    \includegraphics[width=\linewidth]{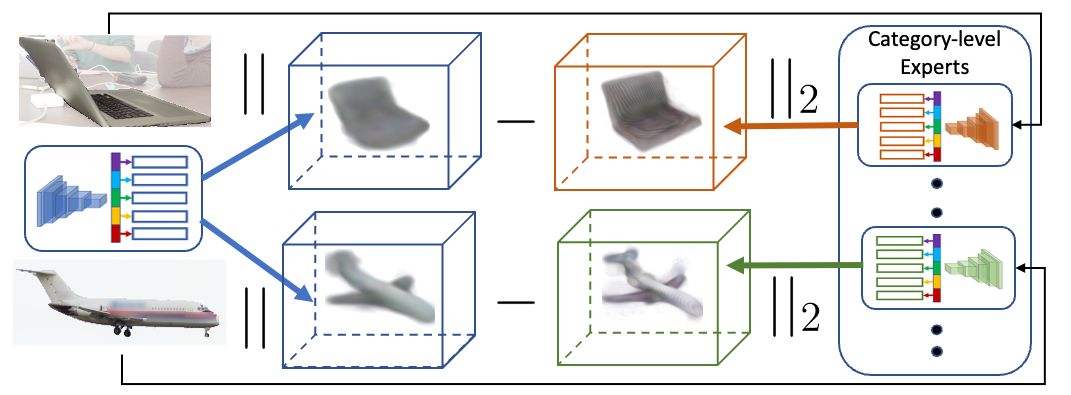}
    \caption{\textbf{Distillation.} We use a point-wise regression loss to distill multiple experts into a unified reconstruction model.}
    \figlabel{distill}
\end{figure}

\subsection{Multi-category Distillation} \seclabel{distillation_approach}
The approach described in \secref{singlecat_approach} allows learning category-specific reconstruction networks from in-the-wild image collections. Denoting by $f^c \equiv (f^c_e, f^c_d)$ a category-specific reconstruction network  learned for a category $c$, we now aim to learn a unified reconstruction network $\bar{f} \equiv (\bar{f}_e, \bar{f}_d)$ that can mimic the learned category-specific `experts' across all categories $c \in \mathcal{C}$. Not only would such a unified network make inference easier, it could also improve performance by leveraging the common structure across classes, and perform better than individual networks, while also allowing \emph{zero-shot} reconstruction for unseen classes.

We propose a simple approach of distilling~\cite{hinton2015distilling} the per-category networks to learn a unified network (see \figref{distill}). Concretely, given the category-level image collections $\mathcal{D}^c$, we train our unified network to match the point-wise density and color predictions from the corresponding category-level experts across all images.
\begin{equation*}
\label{eq:distillation_loss}
  L_{\text{dist}} = \sum_{{c \in C}}~\sum_{{I \in \mathcal{D}^c}}~ \sum_{\mathbf{x}} \| f^c_d(\mathbf{x}, f^c_e(I)) - \bar{f}_d(\mathbf{x}, \bar{f}_e(I)) \|^2_2 
\end{equation*}
While we find it beneficial to sample the points $\mathbf{x}$ along rays, the pointwise distillation objective allows efficient training without any computationally expensive volume rendering steps to aggregate the predictions.





\section{Experiments}
\begin{figure*}[t!]
    \centering
    \includegraphics[width=\textwidth]{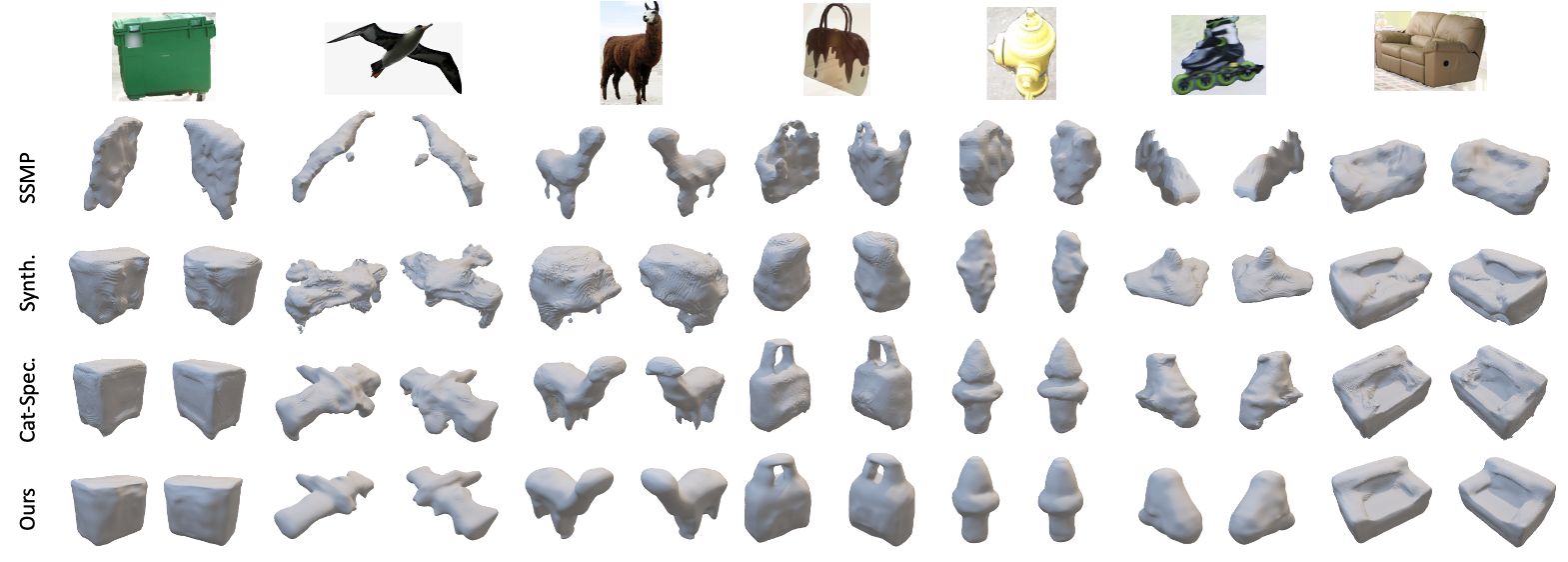}
    \vspace{-2mm}
    \caption{\textbf{Comparisons.} Visualization of 3D reconstructions predicted by the baselines and our final unified model on  OpenImages, CUB, quadrapeds, Co3D and Pascal3D+. The first, second, third and fourth row of visualizations corresponds to predictions of \shelfbl, synthetic data-only model, category-level expert and our final unified model respectively. }
    \label{fig:comparisons}
    \vspace{-3mm}
\end{figure*}

\subsection{Datasets}
\paragraph{Synthetic-data.}
We use a curated collection of CAD models from \cite{3dwarehouse}, spanning 51 object categories and a total of 40k instances used for training.  We render color and mask images for each CAD model from 20 views using Blender \cite{blender}, with azimuth and elevation angles randomly sampled from $[0^\circ,360^\circ]$ and $[-75^\circ,75^\circ]$ respectively.

\vspace{-2mm}
\paragraph{In-the-wild Image Collections.}
As our approach in \secref{singlecat_approach} can learn from image collections with approximate foreground masks, we are able to leverage various image recognition datasets for training. In particular, we use images from the following five datasets:



\noindent \emph{CUB-200-2011\cite{WahCUB_200_2011}}: This consists of 6,000 images with annotated foreground masks from over 200 bird sub-species. We treat the entire image collection as representing a single `bird' category.

\noindent \emph{Quadrupeds from ImageNet}: We use images spanning 25+ quadruped categories from the ImageNet dataset~\cite{deng2009imagenet}, along with approximate foreground masks extracted from an instance segmentation system~\cite{kirillov2020pointrend}. In particular, we use the curated splits from Kulkarni \etal~\cite{kulkarni2020articulation} where they discard images with severe occlusion and truncation, resulting in a total of 25k+ images.

\noindent \emph{PASCAL3D+~\cite{xiang2014beyond}}:  We use the unoccluded instances spanning 11 rigid-categories, with about 3k images per category. We use an off-the-shelf image segmentation system \cite{kirillov2020pointrend} to extract foreground masks for the ImageNet subset of images where ground-truth masks are not available. While this dataset provides annotations for approximate 3D shape in the form of template CAD models, we do not use these during training, and only use these for quantitative evaluation on held-out test images. 

\noindent \emph{Open Images \cite{kuznetsova2020open}}: We use images from across 77 diverse categories, with 500 to 20k images with annotated foreground masks for each category. We adapt the train and test splits from Ye \emph{et al.}~\cite{ye2021shelf}, where a simple image classifier was used to filter out truncated and occluded instances. 

\subsection{Evaluation Setup}

\paragraph{Training Details.}
All our networks share the same architecture. The encoder is a ResNet-34~\cite{he2016deep} with an additional residual linear layer that produces $\mathbf{z} \in \mathbb{R}^{2560}$. Conditioned on $\mathbf{z}$, the decoder is a 5-layer FiLM-ed SIREN network~\cite{chan2021pi} with an embedding size of 256 in each layer. We use Adam optimizer~\cite{kingma2014adam} with a Reduce-on-plateau learning rate scheduler. To ensure better convergence, while learning category-specific models in \secref{synthetic_approach}, we freeze the network for the first 10 epochs wherein we only optimize the multi-hypothesis camera parameters. For additional training details please refer to the supplementary material. 

\vspace{-2mm}
\paragraph{Baselines.} We compare our final  model with \shelfbl, the current state of the art method under similar data assumption as ours \ie learning 3D for generic classes using image collections with approximate foreground masks. We also report the performance of intermediate stages of our approach -- a) the synthetic data-only model (\syntheticbl), and ii) category-specific experts (\singlebl).

\vspace{-2mm}
\paragraph{Evaluation.}
To convert our neural implicit representation $f_d(\cdot, \mathbf{z})$, into an explicit mesh, we follow the Marching Cubes approach suggested by Mescheder \etal~\cite{mescheder2019occupancy}. All our quantitative and qualitative evaluations are reported on this explicit representation. As we have access to (approximate) 3D ground truth meshes in Pascal3D+, we report intersection over union (IoU) on a held out test split. Following CMR~\cite{kanazawa2018learning}, while computing IoU, we search over scale, rotate and position to align the predicted meshes with the ground truth. We also evaluate all approaches on Co3D~\cite{reizenstein2021common}, in which we have access to 3D ground truth point clouds. We evaluate on manually selected 100 occlusion and truncation free instances spanning 10 categories. We report Chamfer distance~\cite{fan2017point} between ground truth point cloud and points uniformly sampled from the predicted surfaces. We follow a similar approach to Pascal3D+ evaluation to align the predictions with ground truth point clouds.

\begin{table}\centering
\scalebox{0.9}{
\begin{small}
\begin{tabular}{@{}lrrrr@{}}\toprule
 & \textbf{\shelfbl} & \textbf{\syntheticbl} & \textbf{\singlebl} & \textbf{\distillbl} \\ \midrule

Aeroplane         &  0.38 &  0.44 & 0.32  & \textbf{0.49} \\
Bicycle       &  \textbf{0.22} &  0.11 &  0.18 & 0.18  \\
Bottle        &  0.43 &  0.54 & \textbf{0.66}  & 0.65  \\
Bus           &  0.21 &  0.42 & 0.53  & \textbf{0.57}  \\

Car           & 0.46  &  \textbf{0.72} & 0.54   & 0.70  \\

Chair     &  0.44 &  0.52 & 0.57  & \textbf{0.59}  \\
D.Table       &  0.26 & 0.28  & 0.31  & \textbf{0.34}  \\
Motorbike     &  0.47 & 0.47  & 0.44  & \textbf{0.53}  \\
Sofa          &  0.24 & 0.43  & 0.46  & \textbf{0.52}  \\
Train         &   0.09 & 0.25 & 0.31  & \textbf{0.37}  \\
TV            &   0.17&  0.19 & 0.27  &  \textbf{0.33} \\
\midrule
\textbf{Mean}          &  0.31 & 0.40  & 0.42  & \textbf{0.48}  \\
\bottomrule
\end{tabular}
\end{small}}

\caption{IoU metric comparison on Pascal3D+~\cite{xiang2014beyond}. Higher IoU implies superior 3D reconstruction.}
\label{table:iou}
\end{table}

\begin{table}\centering
\scalebox{0.9}{
\begin{small}
\begin{tabular}{@{}lrrrr@{}}\toprule
 & \textbf{\shelfbl} & \textbf{\syntheticbl} & \textbf{\singlebl} & \textbf{\distillbl} \\ \midrule

Banana        & 0.256  & 0.409  & 0.265  & \textbf{0.224}  \\
Bottle        & 0.239  & 0.182  & \textbf{0.163} & 0.186  \\
C.Phone       & 0.208  & 0.251  & 0.184  & \textbf{0.172}  \\
Donut         & 0.183  & 0.440  & \textbf{0.176}  & 0.181  \\
Hydrant       & 0.184  & 0.198  &  0.182 & \textbf{0.166}  \\
Orange        & 0.103  & 0.307  & 0.089  & \textbf{0.082}  \\
Suitcase      & 0.162  & 0.310  & \textbf{0.146}  & 0.149  \\
T.Bear        & \textbf{0.197}  & 0.427  &  0.239 & 0.245  \\
Toaster       & 0.194  & 0.316  & 0.161  & \textbf{0.153}  \\
Vase          & 0.210  & 0.286  & 0.211  & \textbf{0.197}  \\

\midrule
\textbf{Mean}          &  0.194 & 0.313  & 0.182  & \textbf{0.176}  \\
\bottomrule
\end{tabular}
\end{small}}

\caption{Chamfer distance comparison on Co3D~\cite{reizenstein2021common}. Lower Chamfer distance implies superior 3D reconstruction.}
\vspace{-5mm}
\label{table:chamfer}
\end{table}

\subsection{Results}

\paragraph{Quantitative Results.}
We report empirical results for Pascal3D+ and Co3D dataset in Tables.~\ref{table:iou} and~\ref{table:chamfer} respectively, and note that our final model consistently improves over the prior state-of-the-art as well as the synthetic baseline. We also observe that the joint model yields some overall improvements over the per-category experts.

\begin{figure}[t!]
    \centering
    \includegraphics[width=\linewidth]{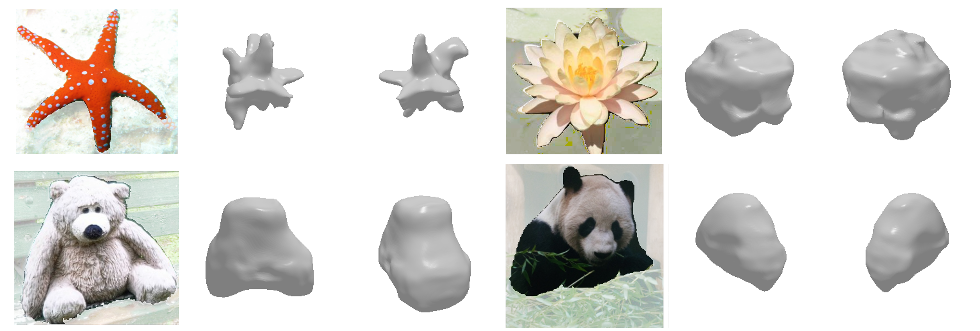}
    \caption{\textbf{Failure modes.} Our model is not robust to occlusion and does not perform well on some difficult categories.}
    \label{fig:failures}
\end{figure}

\begin{figure*}[!t]
    \centering
    \includegraphics[width=0.99\textwidth, trim={0 0 0 10cm},clip]{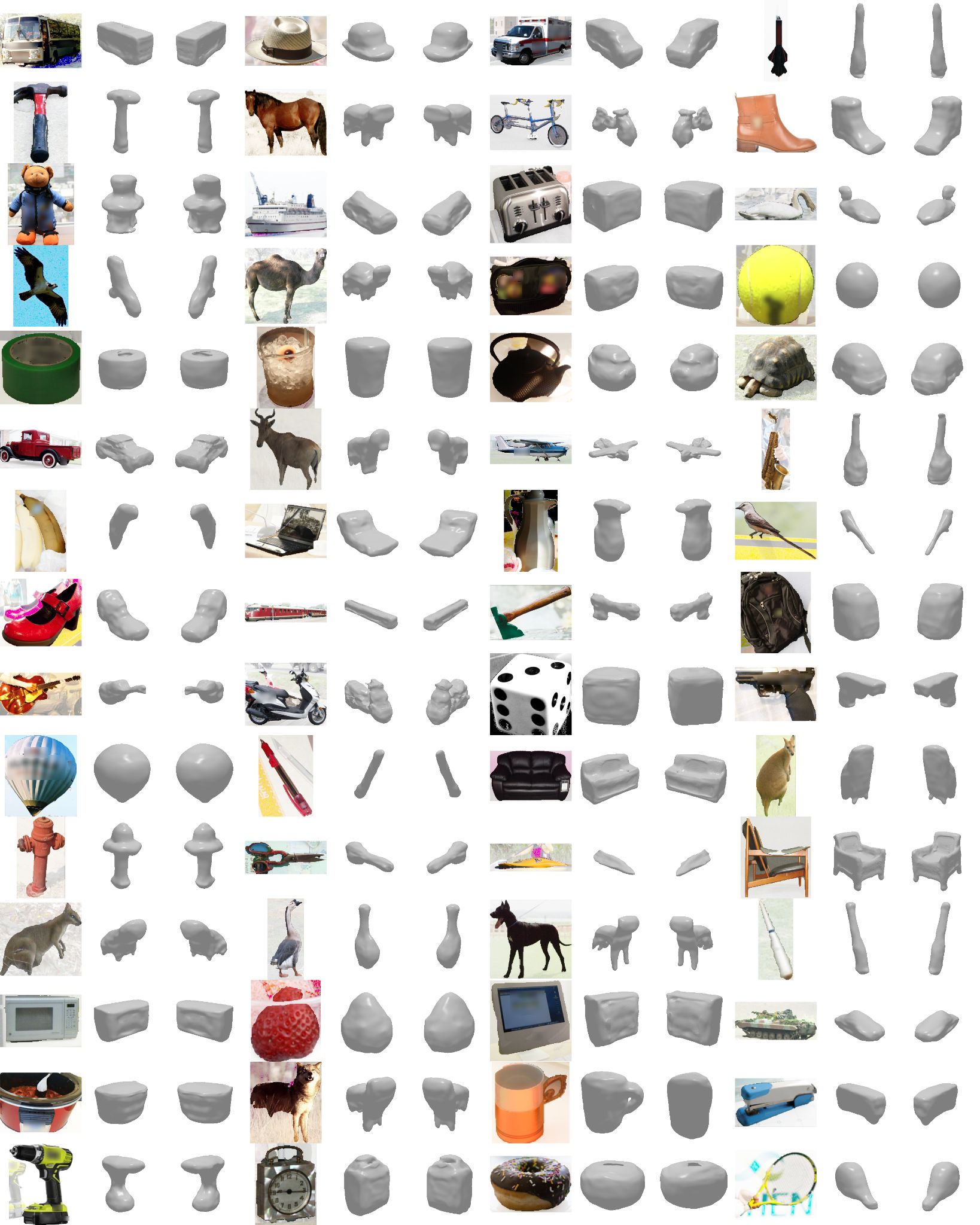}
    \caption{\textbf{In-the-wild Results.} Single-view 3D inference of our final unified model in the wild. We visualize from two viewpoints the meshes extracted from the predicted neural representation in a canonical frame.}
    \label{fig:wild}
\end{figure*}

\vspace{-2mm}
\paragraph{Qualitative Evaluation.} We demonstrate the 3D reconstruction quality of our final unified model in a wide range of settings in Fig.~\ref{fig:teaser},~\ref{fig:wild}. We also visualize qualitative comparisons with baselines in Fig.~\ref{fig:comparisons}, and find that these corroborate the quantitative analysis -- that our unified model improves over the baselines and is capable of producing high-quality reconstructions for a wide variety of classes. We also depict some representative failure modes in Fig. \ref{fig:failures}. 

\vspace{-2mm}
\paragraph{Zero-shot Reconstruction.}
A benefit unique to learning a unified 3D reconstruction model is that, because of the common structure across generic objects, it can perform meaningful reconstruction even on \emph{unseen} categories. We evaluate this ability of the final unified model to perform zero-shot reconstruction  by varying the number of category-specific experts available to distill. We consider two unified models (\distillbl-75+) and (\distillbl-125+), which are distilled from over 75 and 125 experts respectively, but crucially, have not train on the evaluation categories. We summarize our observations in Table.~\ref{table:zero_shot}, and find a clear and encouraging trend that observing more classes in training improves performance on \emph{unseen} categories.

\begin{table}\centering
\scalebox{0.9}{
\begin{small}
\begin{tabular}{@{}lrrrr@{}}\toprule
  & \textbf{\syntheticbl} &\textbf{\distillbl}-75+ & \textbf{\distillbl}-125+  \\ \midrule
Banana    & 0.409  & 0.378  &  \textbf{0.304} \\
Donut     & 0.440  & 0.412  & \textbf{0.244}   \\
Orange    & 0.307  & 0.218  &  \textbf{0.213}  \\
Suitcase  & 0.310  & 0.286  &  \textbf{0.189} \\
T.Bear    & 0.427  & 0.364 &  \textbf{0.320}  \\
Toaster   & 0.316  & 0.194 &  \textbf{0.167}  \\

\midrule
\textbf{Mean}      &  0.368  &  0.308 & \textbf{0.239} \\
\bottomrule
\end{tabular}
\end{small}}

\caption{\textbf{Zero-shot reconstruction on unseen classes}. We report the Chamfer loss on unseen classes in Co3D dataset. As the model is trained over more classes, its zero-shot reconstruction performance on held-out unseen object classes improves.}
\label{table:zero_shot}
\vspace{-5mm}
\end{table}

\section{Conclusion}

We presented a simple and scalable approach to learn a unified 3D reconstruction model across a diverse set of object categories. We achieved this by pre-training on synthetic data, which then allowed self-training on a variety of image collections. This simple strategy enabled us to learn a state-of-the art reconstruction model that can infer 3D for over hundred categories, and our results also highlight the empirical benefits of this sharing. While these results are encouraging, we note that our approach has some common failure modes. In particular, our reconstructions are unable to capture fine shape details. Our approach also crucially relies on images of unoccluded isolated objects for learning, and thus can neither learn from, nor infer 3D in more challenging images.



{\small
\bibliographystyle{ieee_fullname}
\bibliography{main}
}

\clearpage
\section*{A1. Additional Training Details}

\paragraph{Input.} In all phases of training, we consider zero-padded single-instance only color images as input to our network. Particularly, the input to our network is a square image of size 224 where the larger dimension of the instance bounding box is resized to 224. The images correspond to segmented objects, and the background pixel values are also set to zero.

\paragraph{Pre-Training from Synthetic 3D Data. }
We consider multi-view supervision for training training the model. For each input image in each iteration, we consider rays from 5 other views (among a total of 20 available views) and their associated cameras for supervision. Unlike input images which have a size of 224, the images used for supervision are resized to 128x128. This is to ensure that we can cover more region with a single ray for faster training of the network. All the camera poses we use for supervision are with respect to a standardised canonical object frame with X axis being left to right, and Z axis being upright. From each label instance, we sample 340 rays for supervision. Thus, the effective number of rays we use for supervising each input image is 5*340. Additionally, on each camera ray we query 100 points uniformly up to 2 meters during volume rendering. Thus for each input instance we query 5*340*100 number of points from the input-conditioned decoder. 

We initialize our ResNet-34 based encoder with pre-trained weights from ImageNet-1k classification task. These pre-trained weights are borrowed from TorchVision. And our decoder's weights are initialized uniformly using PyTorch's default initialization scheme. We train our network for about 450 epochs in multi-gpu setting using Distributed Data Parallel (DDP). We train on 32 Nvidia-V100 32GB GPU's with a batch size of 8 on each GPU. Thus, our effective batch size is 8*32. This approximately took about 2.5 days to train using 32 GPU's. We train our network with Adam optimizer \cite{kingma2014adam} with a starting learning rate of $5^{-5}$ that is updated conditionally using a reduce on plateau learning rate scheduler with a reduction factor and threshold of $0.5$ and $5^{-3}$ respectively.

\paragraph{Self-Training from Image Collections.}
As mentioned in the main manuscript, we don't have access to cameras during this phase and we train our models to infer shape in canonical object frame using multi-hypothesis cameras. At the beginning of training, for each instance, we initialize a multi-hypothesis camera consisting of 10 cameras randomly. For the first 10 epochs, we freeze the model weights and update each multi-hypothesis camera using the multi-hypothesis camera loss suggested in the original manuscript. This  ensures that even before we begin training our network, our multi-hypothesis cameras are at an acceptable pose. We then learn both our model and  multi-hypothesis cameras for additional 40 epochs. Another important point to note is that we have a different update rate for multi-hypothesis cameras and model parameters. In each mini-batch step, we update multi-hypothesis camera parameters 10 times for every model update. This is achieved through an additional multi-hypothesis camera optimization loop in the training step. 

We consider Adam optimizer with a constant learning rate of $10^-6$ and  $0.1$ for training for model and multi-hypothesis cameras respectively. We again train each category-specific model on 16 Nvidia-V100 32 GPU's using DDP training strategy with a batch size of 4 on each GPU. This phase took about 1-3 days for each category-level expert based on the dataset size.

Additionally, similar to pre-training, for each camera in multi-hypothesis camera we consider 340 rays with 100 samples on each ray.

\paragraph{Multi-Category Distillation.}
Unlike prior phases that rely on ray-wise supervision, we perform a point-wise supervision to train our final unified model. We randomly sample 25000 3D points in the standardized canonical object frame for every image in each step step.

We train this network using Adam optimizer with a learning rage of $10^{-6}$ for 75 epochs on 32 Nvidia-V100 GPU's using DDP. This training phase is relatively faster in comparison to prior phases and took about 2 days to distill all category-level experts and synthetic data-only model into a single model.

\end{document}